\documentclass{article}

\usepackage[english]{babel}

\usepackage[letterpaper,top=2cm,bottom=2cm,left=3cm,right=3cm,marginparwidth=1.75cm]{geometry}

\usepackage{multirow}
\usepackage{amsmath,amssymb,amsthm}
\theoremstyle{definition} 
\usepackage{relsize}
\usepackage{graphicx}
\usepackage{listings}
\usepackage{xcolor}
\usepackage{booktabs}
\usepackage{fancyref}
\usepackage[parfill]{parskip}
\usepackage{todonotes}
\usepackage{import}
\usepackage{dsfont}
\usepackage{wrapfig} 
\usepackage{amsmath}
\usepackage{array}
\usepackage{booktabs}

\usepackage{lineno}
\usepackage{float}
\usepackage{caption}
\usepackage[skip=10pt]{subcaption}
\usepackage{bm}
\usepackage{bbm}
\usepackage{commath}
\usepackage[section]{placeins}
\usepackage{fancyhdr}
\usepackage[style=authoryear,backend=biber,giveninits=true,natbib=true,maxcitenames=2,mincitenames=1,maxbibnames=3]{biblatex}
\usepackage{datetime}
\usepackage{float}
\usepackage{algorithm}
\usepackage{algpseudocode}
\mathchardef\mhyphen="2D % Define a "math hyphen"

\newcommand{\indep}{\perp \!\!\! \perp} % independence symbol
\newcommand{\notindep}{\not\!\perp \!\!\! \perp} % not independence symbol
\numberwithin{equation}{section}

\usepackage{amsmath}
\usepackage{mathtools}

\theoremstyle{plain} \newtheorem{theorem}{Theorem}[section]
\usepackage{graphicx}
\usepackage[colorlinks=true, allcolors=blue]{hyperref}

\newcommand{\myref}[2]{\hyperref[#2]{#1 \ref*{#2}}}

\title{Efficient Causal Discovery for Autoregressive Time Series}
\author{
    Mohammad Fesanghary \\
    Bloomberg \\
    New York, NY, USA \\ 
    \texttt{mfesanghary1@bloomberg.net}\\ 
\and
    Achintya Gopal \\
    New York, NY, USA \\ 
    \texttt{achintyagopal@gmail.com} 
}
\begin{document}
\maketitle

\begin{abstract}
In this study, we present a novel constraint-based algorithm for causal structure learning specifically designed for nonlinear autoregressive time series. Our algorithm significantly reduces computational complexity compared to existing methods, making it more efficient and scalable to larger problems. We rigorously evaluate its performance on synthetic datasets, demonstrating that our algorithm not only outperforms current techniques, but also excels in scenarios with limited data availability. These results highlight its potential for practical applications in fields requiring efficient and accurate causal inference from nonlinear time series data.

\end{abstract}

\section{Introduction}
Causal structure learning (CSL) in time series refers to the process of identifying and quantifying potentially time-lagged causal relationships among variables in a system. Unlike traditional time series analysis, which often focuses on prediction and correlation, CSL aims to uncover the cause-and-effect relationships that underlie the observed data. CSL is a crucial challenge in numerous fields such as economics, finance, healthcare, and natural science, where understanding the causal mechanisms can lead to more accurate forecasting, targeted interventions, and improved risk management.

Causal structure learning poses significant challenges due to the presence of unobserved confounding factors, limited observational data, non-stationarity, and noise. Traditional CSL methods, which primarily focus on contemporaneous data, address some of these issues, but encounter considerable difficulties when extended to time series data. This complexity arises because time series CSL requires incorporating an appropriate number of lagged variables to capture temporal dependencies, resulting in a substantially larger model. Consequently, fitting such an expanded model demands significantly more computational resources, making these methods less efficient and scalable for time series analysis. 

Work on time series causal discovery can be categorized into score-based and constraint-based methods. The former attempt find relationships that optimize a ``score'' such as the Bayesian Information Criterion (BIC) or Akaike Information Criterion (AIC) \citep{chickering2002optimal, scutari2010learning}, while the latter employ several conditional independence tests to establish and/or disprove plausible causal relationships \citep{kalisch2007estimating}. Both methods have distinct advantages and drawbacks. Score-based methods are flexible, can handle large networks well, and simultaneously perform structure learning and parameter estimation, but they can be less interpretable, computationally intensive, and prone to overfitting. Conversely, constraint-based methods are more interpretable, grounded in conditional independence theory, and require fewer assumptions, but they rely heavily on the reliability of statistical tests, and can struggle with large networks and datasets due to the combinatorial growth of tests. 

In this work, we address the computational complexity of the aforementioned constraint-based algorithms by decreasing the required number of tests to grow quadratically rather than exponentially. To achieve this goal, we build upon the SYPI algorithm  \citep{mastakouri2021necessary}, which was originally introduced for causal feature selection in linear time series data. We extend the SYPI algorithm to handle nonlinear time series data, such as asset price dynamics or climate patterns, and incorporate a pruning step to control false positive links identified during the discovery process. 
In \myref{Section}{sec:methodology}, we describe the algorithm and its assumptions. In \myref{Section}{sec:experiments}, we test its performance against other time series causal discovery methods. Finally, in \myref{Section}{sec:case_study}, we conclude with a case study  analyzing the causal connections between major banks.

\section{Methodology}\label{sec:methodology}

As previously mentioned, this work extends the SyPI method  \citep{mastakouri2021necessary}. Before introducing our algorithm, SyPI+, we first revisit the original SyPI framework, outlining its key assumptions and methodological details to provide a clear foundation for the extensions that follow. 

SyPI is a causal feature selection approach for time series that effectively identifies both direct and indirect causes from observational data, while being robust to a specific class of latent confounders, namely those that directly affect the target variable, but not other observed variables (see A9 in \myref{Table}{table:assumptions}). 

SyPI constructs the conditioning set directly, eliminating the need for exhaustive searches. This direct construction reduces the number of conditional independence (CI) tests to just two per candidate time series, significantly decreasing the overall computational burden and mitigating the issues associated with multiple testing. 

%\subsection{Assumptions}
In order to be able to build the condition set, SyPI needs a set of assumptions and conditions (see \myref{Table}{table:assumptions}) that ensure soundness and completeness. Assumptions A1-A5 are standard in constraint-based causal discovery methods. Additional assumptions A7-A9 impose some restrictions on the connectivity of the graph, and assume all variables have one lag dependency to their own past. Assumption A6 states that the target time series has no children. In this work, we relax this assumption and instead introduce a pruning step at the end of the pipeline to remove edges that may have been introduced due to its violation.

\begin{table}[h!]
\caption{Assumptions for SyPI method}
\centering
\begin{tabular}{@{}c p{12cm}@{}}
\toprule
\textbf{Assumption} & \textbf{Description} \\ 
\midrule
\textbf{A1}         & Causal Markov condition \\ 
\textbf{A2}         & Causal Faithfulness in the full time graph. \\ 
\textbf{A3}         & No backward arrows in time \(X^{i}_{t^\prime} \not\to X^{j}_{t}, \forall t^\prime > t\). \\ 
\textbf{A4}         & Stationary graph: the graph is invariant under a joint time shift of all variables. \\ 
\textbf{A5}         & The graph is acyclic. \\ 
\textbf{A6}         & The target time series \(Y\) is a sink node. (we will relax this). \\ 
\textbf{A7}         & There is an arrow \(X^{i}_{t-1} \rightarrow X^{i}_{t}, Y_{t-1} \rightarrow Y_t, \forall i, t \in \mathbb{Z}\). \\ 
\textbf{A8}         & There are no arrows \(Q^{i}_{t-s} \rightarrow Q^{i}_{t}\) for \(s > 1\). \\ 
\textbf{A9}         & Every variable \(U^i\) (unobserved confounder) that affects \(Y\) directly or that is connected with an observed collider in the summary graph should be memoryless (\(U^{i}_{t-1} \rightarrow U^{i}_{t}\)) and should have single-lag dependencies with \(Y\) in the full time graph. \\ 
\bottomrule
\end{tabular}
\label{table:assumptions}
\end{table}

\subsection{Construction of Condition Set}
In order to build the condition set that is sufficient to identify whether \(X^{j}\) is a cause of \(Y\) we need to find the minimum lag (\(w\)) dependency between \(X\) variables and \(Y\).  \cite{mastakouri2021necessary} defines lag \(w\) for the ordered pair of a time series \(X^{i}\) and the target Y if there exists a collider-free path \(X^{i}_{t}---Y_{t+w}\) that does not contain a link of this form  $Q^{r}_{t} \rightarrow  Q^{r}_{t+1}$, with arbitrary \(t\), 
for any \( r \not\equiv i,j \), nor any duplicate node, and any node in this path does not belong to \(X^{i}\),\(Y\) . See explanatory \myref{Figure}{fig:min_lag}. We will discuss the ways to determine \(w\) in the next section.

\begin{figure}[!bt]
\begin{minipage}{\linewidth}
    \centerline{\includegraphics[width=0.4\linewidth]{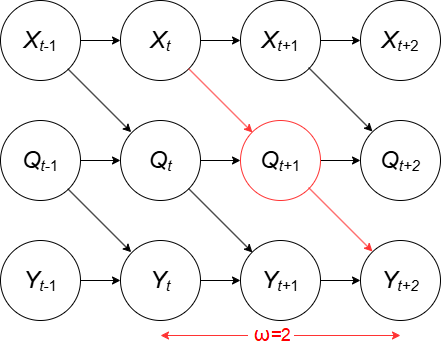}}
    \caption{In this figure, min lag is 2. }\label{fig:min_lag}
\end{minipage}
\end{figure}

\begin{theorem}[Theorem 1 from \cite{mastakouri2021necessary}]
For every \(X^{i} \in \mathbf{X} \), we define a conditioning set as:

 \( \mathbf{S^{i}} = \{ X^{1}_{t+w_{i1}-1}, X^{2}_{t+w_{i2}-1},...,X^{i-1}_{t+w_{i,i-1}-1},X^{i+1}_{t+w_{i,i+1}-1}, ..., X^{n}_{t+w_{in}-1} \} \)
where \(w_{i}\) is the minimum lag between \(X^{i}\) and Y, and \( w_{ij} := w_{i}-w_{j}  \).

If 
\begin{equation}
  X^{i}_{t} \notindep Y_{t+w_{i}} \mid \{\mathbf{S^{i}},  Y_{t+w_{i}-1}   \}  
\end{equation}  
and
\begin{equation}
X^{i}_{t-1} \indep Y_{t+w_{i}} \mid \{\mathbf{S^{i}}, X^{i}_{t}, Y_{t+w_{i}-1} \} 
\end{equation} 
are true, then \( X^{i}_{t}  \rightarrow Y_{t+w_{i}} \).

\end{theorem}

Please see \citet{mastakouri2021necessary} for the proof and detailed discussions. At this stage, we iterate over all variables to assess whether any could be potential causes, based on the two CI tests described above. The pseudocode of our algorithm, SyPI+, is presented in \myref{Algorithm}{alg:pseudo_code}.

\begin{algorithm}[t]
   \caption{SYPI+ Pseudocode}
   \label{alg:pseudo_code}
\begin{algorithmic}
\Require set of variables $\mathbf{X}$
% \State {\bfseries Output:} causal graph
\For{each $Y$ in $\mathbf{X}$}
    \State $\mathbf{R} = \mathbf{X} \setminus \{Y\}$
    \State $w = \text{min\_lags}(\mathbf{R}, Y)$ \Comment{See \myref{Figure}{fig:min_lag}}
    \State $n_{vars} = \abs{\mathbf{R}}$ 
    \For{$i=1$ to $n_{vars}$}
        \State $\mathbf{S}_i = \bigcup_{j=1, j \neq i}^{n_{vars}} \{X^{j}_{t + w_i - w_j - 1}\}$
        \State $\text{p\_value}_1 = \text{ci\_test}(X^i_t, Y_{t + w_i}, [\mathbf{S}_i, Y_{t + w_i - 1}])$ \Comment{$X^i_t \indep Y_{t + w_i}\ |\ \mathbf{S}^i, Y_{t + w_i - 1}$ ?}
        \If{$\text{p\_value}_1 < \text{threshold}_1$}
            \State $\text{p\_value}_2 = \text{ci\_test}(X^i_{t-1}, Y_{t + w_i}, [\mathbf{S}_i, X^i_t, Y_{t + w_i - 1}])$ \Comment{$X^i_{t-1} \indep Y_{t + w_i}\ |\ \mathbf{S}^i,  X^i_t, Y_{t + w_i - 1}$ ?}
            \If{$\text{p\_value}_2 > \text{threshold}_2$}
                \State \text{Add $X^i_t$ as cause}
            \EndIf
        \EndIf
    \EndFor
\EndFor
\State \text{Prune Graph}
\State \Return 
\end{algorithmic}
\end{algorithm}

In the next two sections we will focus on possible ways to identify minimum lags and discuss pruning logics that our algorithm performs in order to discover false positive links between our variables.

\begin{figure}[!bt]
\begin{minipage}{\linewidth}
    \centerline{\includegraphics[width=0.70\linewidth]{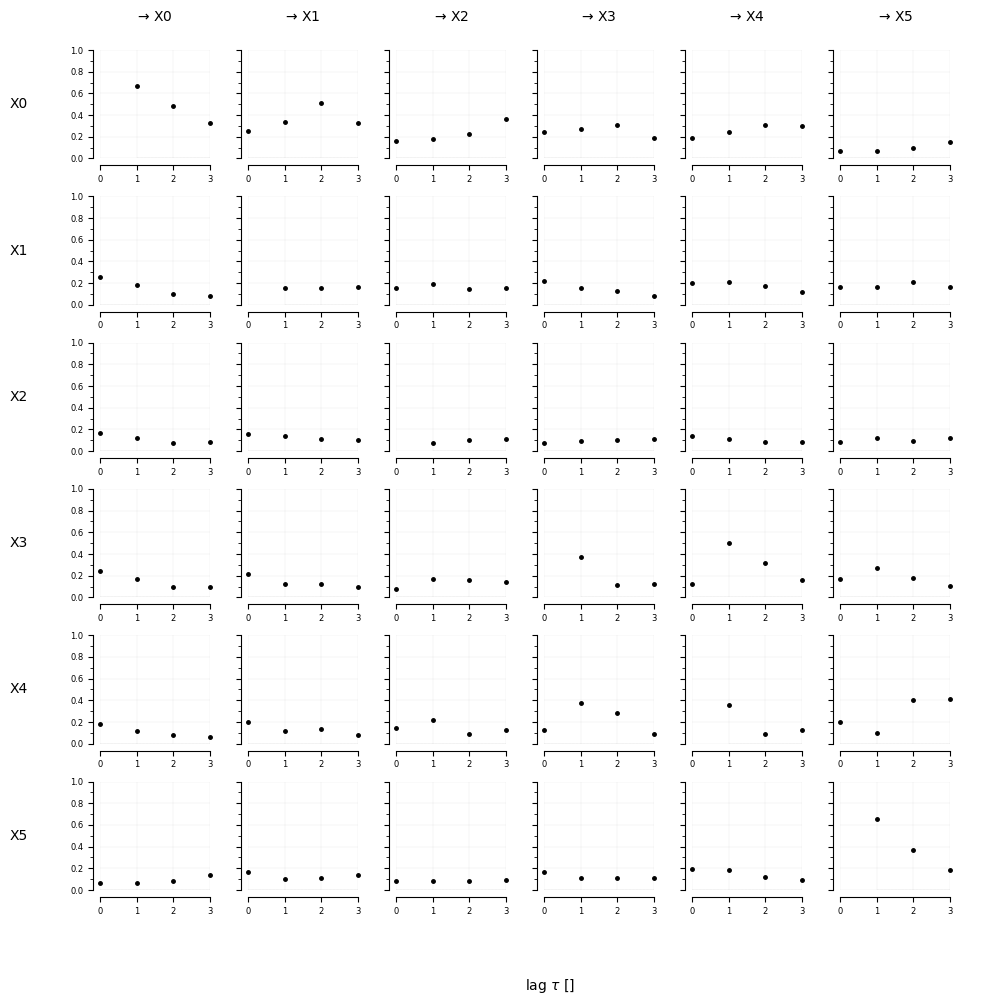}}
    \caption{Lag Graph: Each row represents a candidate cause variable, and each column corresponds to target variable at time t. The values indicate the dependency (e.g., Pearson correlation, distance correlation, mutual information) of various lags of the row variable on the column variable at time t.}\label{fig:lag_graph_example}
\end{minipage}
\end{figure}

\subsection{Minimum-lag (\(w\)) Selection}
The process of identifying minimum lag between two time series, as defined above, is very important for ensuring correct causal connections. In the original work, \citet{mastakouri2021necessary} recommend using Lasso regression; however, since we are dealing with nonlinear connections, we cannot use lasso regression to identify the minimum lag. The strategy we propose involves generating lag plots and computing nonlinear dependency metrics, such as distance correlation, multiscale graph correlation, or conditional mutual information. In this work, we utilize distance correlation in the lag plots to effectively capture potential nonlinear dependencies between variables across time.

We generate lag plots for the time series data using all lags from 0 up to a specified maximum lag T (see  \myref{Figure}{fig:lag_graph_example} for example). A lag plot is a scatter plot that visualizes the relationship between the current value of a time series and its own lagged values or those of other time series. This visualization aids in identifying dependencies across time and can also inform the selection of an appropriate maximum lag for the problem at hand. For instance, in the lag plot shown in \myref{Figure}{fig:lag_graph_example}, examining the second column reveals that the only potential cause appears to be  \( X^{0}_{t-2}\). Additionally, by inspecting the second row from the top, it becomes evident that none of the lags of \(X_{1}\) show significant dependency on any other variable. 

After computing the dependency metrics for all lags up to a maximum lag T, we identify the lags exhibiting the highest dependencies. A threshold of 0.05 is applied to the distance correlation values to filter out weak associations. In practice, however, the differences in dependency values across lags may not be significant—this can result from noise in the data or intrinsic properties of the time series. In such cases, where multiple lags exhibit similarly high dependency values, we retain all of them to avoid excluding the true causal lag.

\subsection{Pruning}

In causal discovery, it is essential to ensure that the identified causal links genuinely represent underlying causal relationships and are not artifacts or spurious connections. Pruning is a critical step in refining the causal graph by removing such spurious links. The process involves systematically evaluating and removing links that do not have strong support from the data. Here is a detailed explanation of the approach focusing on simple cycles and common ancestors (see  \myref{Figure}{fig:pruning} for example):

\emph{Simple Cycles}:
    A simple cycle in a graph is a path that starts and ends at the same node without repeating any nodes or edges along the way.
Cycles in a causal graph can indicate potential redundancy or misidentified causal directions. For each edge in a simple cycle, we test if its removal makes the graph more consistent with the conditional independencies in the data. The condition set that we use is the Markov blanket of the nodes. If the data supports the removal (i.e., if the conditional independencies hold better without the edge), the edge is pruned.

\emph{Common Ancestors}:
    A common ancestor is a node that has directed paths leading to two or more nodes in the graph.
In the presence of a common ancestor, direct connections (edges) between its descendants may be spurious because the observed dependency could be due to the common ancestor rather than a direct causal link. In this scenario for each direct link between nodes that share a common ancestor, we test if removing the link aligns better with the conditional independencies implied by the data. If removing the link is more consistent with the data, the link is pruned.

\begin{figure}[!bt]
\begin{minipage}{\linewidth}
    \centerline{\includegraphics[width=\linewidth]{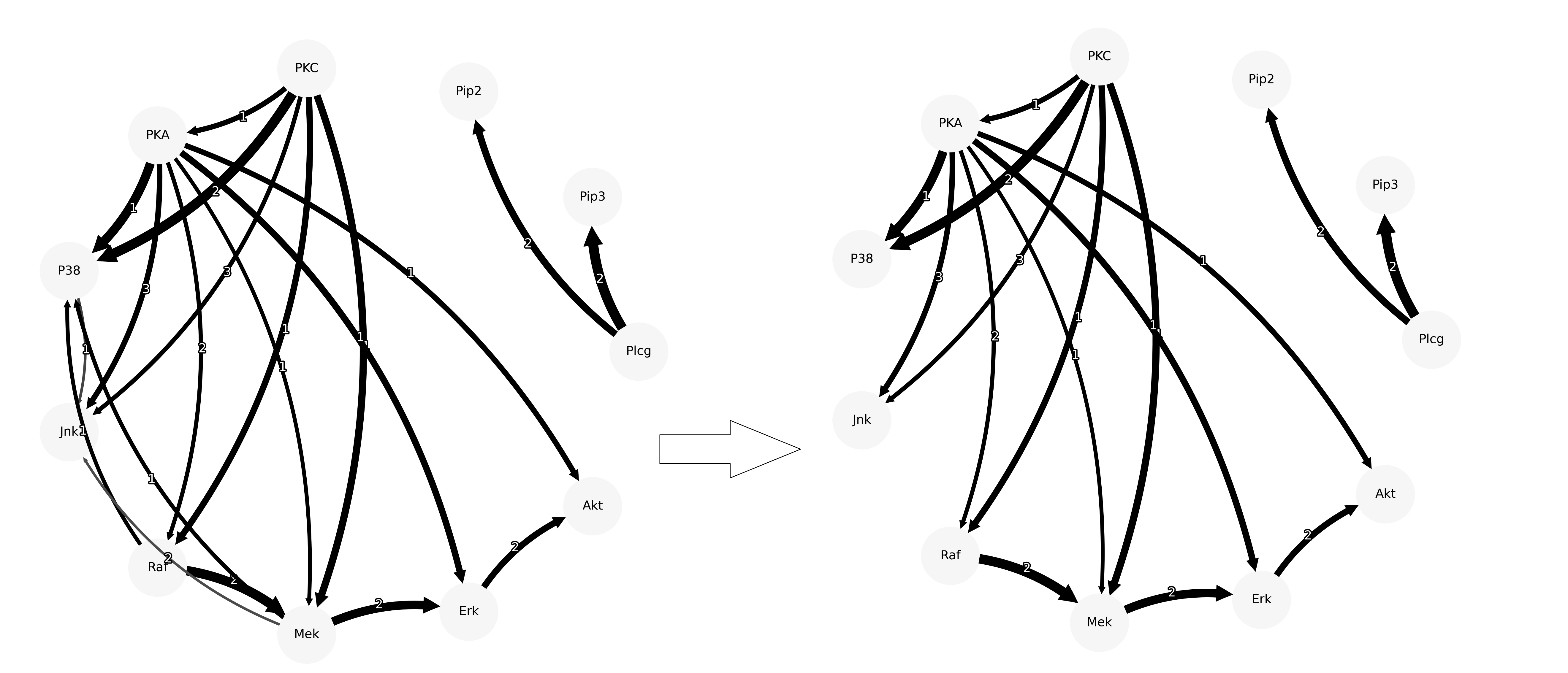}}
    \caption{Pruning removes spurious links. All direct connections between Mek, Raf, Jnk, and P38 are spurious and were induced by common ancestors —namely PKA and PKC.}\label{fig:pruning}
\end{minipage}
\end{figure}

\section{Synthetic Experiments}\label{sec:experiments}

The choice of CI test and data availability can greatly affect the performance of constraint-based algorithms. In this section, we are trying to perform extensive tests in order to study the behavior of the algorithm in different scenarios regarding the availability and complexity of data as well as sensitivity to the choice of CI tests.

\subsection{Dataset}
We experimented with graphs containing 3, 4, 5, 6, 8, 10, and 15 nodes, generating a total of 350 random graphs (50 graphs for each node count). For each graph, we created datasets with 50, 150, 300, 500, and 1,000 data points. The relationships between connected nodes varied and could be linear or nonlinear (quadratic, exponential, or sine). The maximum lag allowed was five.

\subsection{CI Tests}
We tested the following CI tests:
\begin{itemize}
    \item \textbf{Partial Correlations (ParCorr)}: Assumes linearity.
    \item \textbf{Kernel-based Conditional Independence Test (KCIT)}
    \item \textbf{Randomized Conditional Correlation Test (RCoT)}
    \item \textbf{Conditional Mutual Information using k-nearest neighbors (CMIknn)}
\end{itemize}

For KCIT and RCoT, we examined p-value computations using the Satterthwaite-Welch (SW) method  \citep{Welch38,Satterthwaite46} or Hall–Buckley–Eagleson (HBE) method \citep{Hall83,Buckley1988}, which approximate the CDF of a weighted sum of squared normals.

\subsection{Results and Observations}
 We compare the results of SyPI+ with those of CD-NOTS \citep{sadeghi2024causal} and PCMCI \citep{runge17} across all test cases, as these algorithms belong to the same family of constraint-based methods for time series and are therefore comparable.
 
 \textbf{Computational Cost}: 
When selecting a conditional independence (CI) test, computational cost is a crucial factor. \myref{Figure}{fig:ci_runtime} illustrates the average runtime of various CI tests across different sample sizes. Generally, methods like CMIKnn are significantly slower than simpler linear tests such as ParCorr, limiting their use in large datasets. KCIT's computational expense is moderate, scaling with $\mathcal{O}(n^3)$ as the sample size increases. In contrast, RCoT has a time complexity of $\mathcal{O}(m^2n)$, where $m$ is the user-defined number of Fourier terms (typically between 10-100), offering notable computational savings over KCIT.

As expected, the number of conditions needing testing grows quickly with the number of nodes, as depicted in  \myref{Figure}{fig:ci_tests}. A major benefit of SyPI+ is its ability to avoid the combinatorial growth of tests as network size increases. It is important to note that the required number of tests for a given network size can vary between CI tests. This variation occurs because early edge removals in skeleton discovery, influenced by the test's p-value, can significantly reduce the number of condition sets needing evaluation, thus affecting the total number of tests.

\begin{figure}[!bt]
\begin{minipage}{\linewidth}
    \centerline{\includegraphics[width=0.55\linewidth]{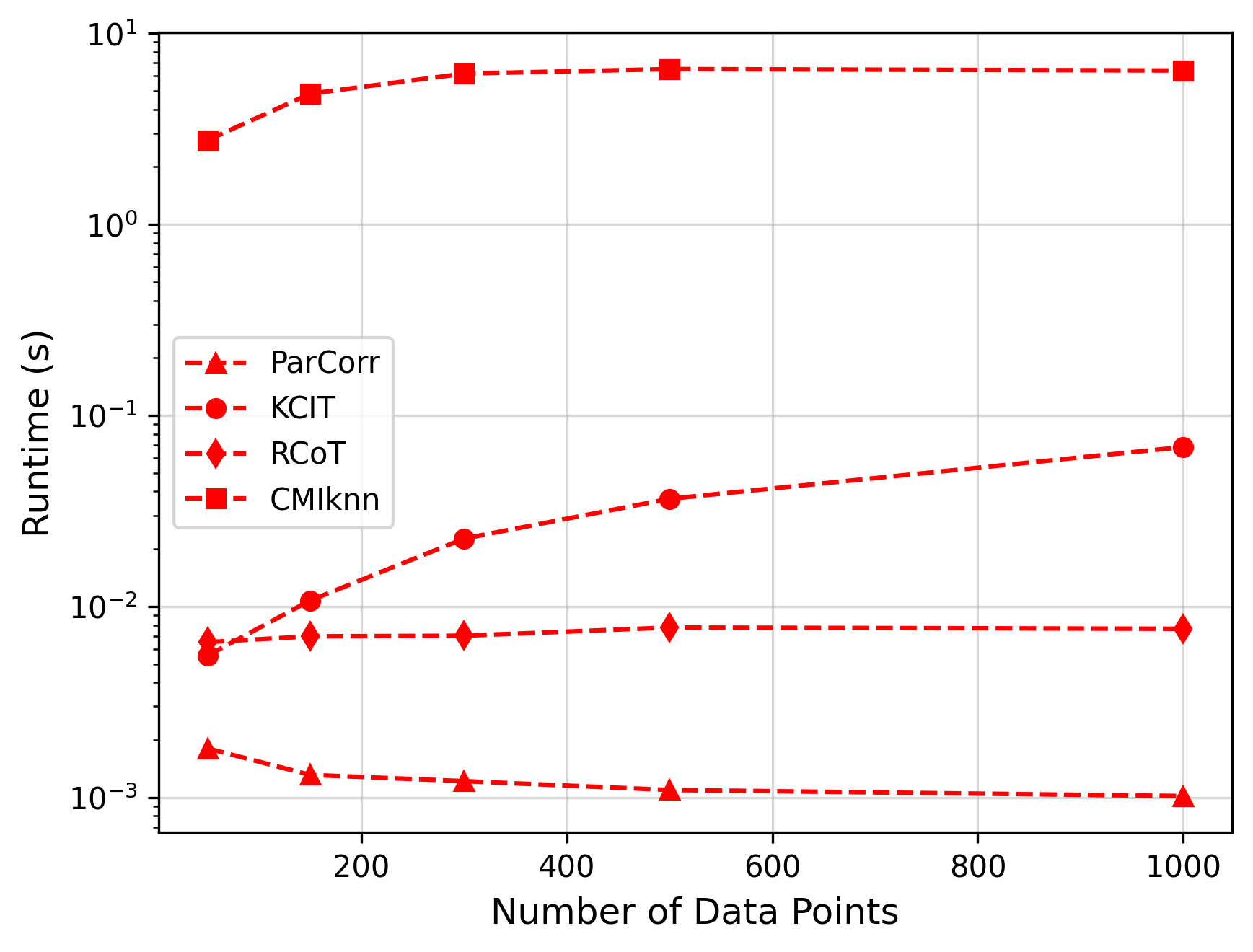}}
    \caption{Runtime evaluation for SyPI+ for different CI tests}\label{fig:ci_runtime}
\end{minipage}
\end{figure}
\begin{figure}[!bt]
\begin{minipage}{\linewidth}
    \centerline{\includegraphics[width=.8\linewidth]{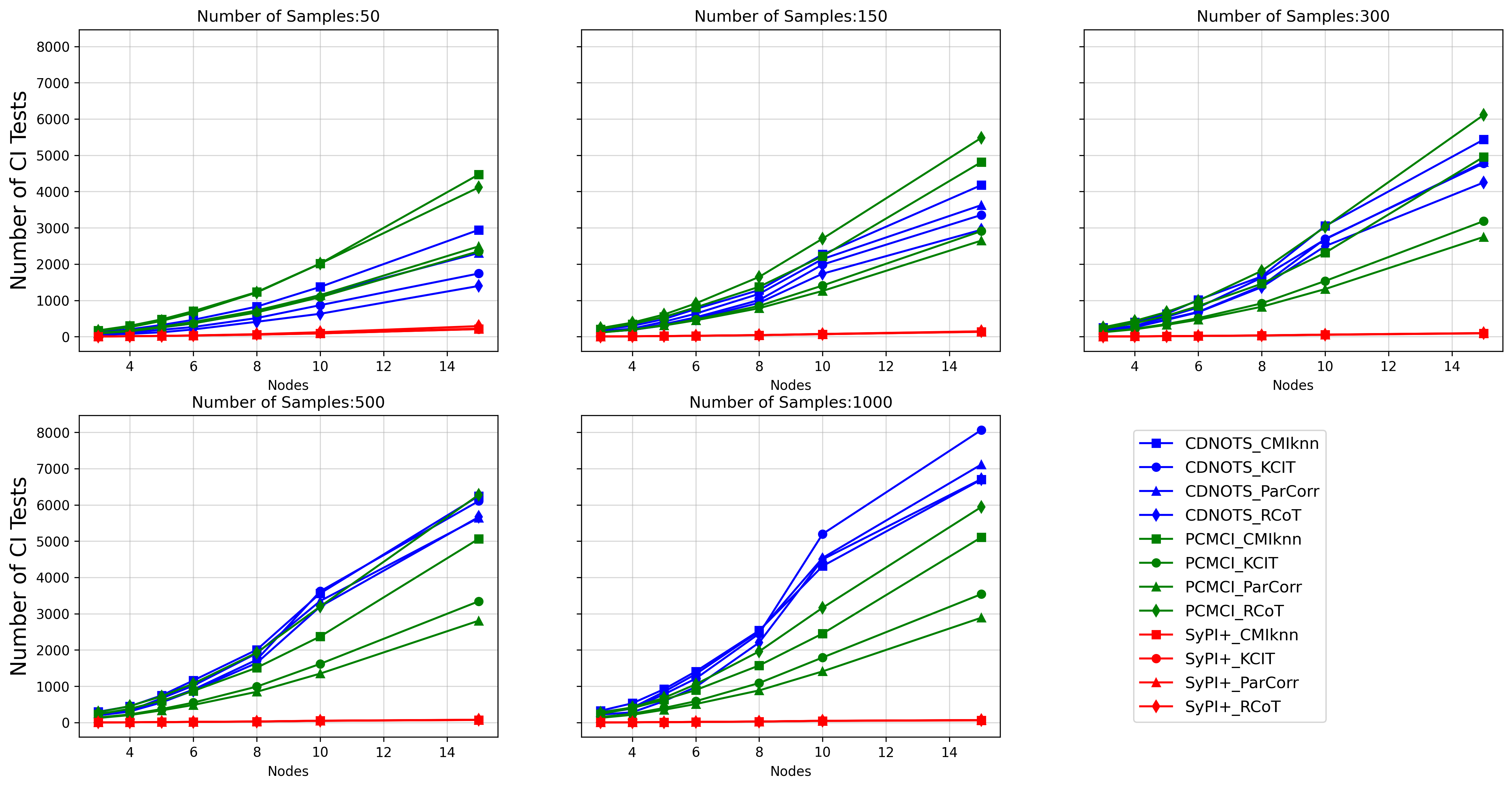}}
    \caption{Number of CI tests executed by each algorithm across different test types}\label{fig:ci_tests}
\end{minipage}
\end{figure}

\textbf{F-Score}: 
\myref{Figure}{fig:ci_fscore_3d} displays the average F-scores\footnote{F-score is the harmonic mean of precision and recall metrics i.e., F-score$=2\,\frac{precision\times recall}{precision+recall}\,$.} for each CI test across different node counts and data points. Key observations include: 

\begin{figure}[!bt]
\begin{minipage}{\linewidth}
    \centerline{\includegraphics[width=\linewidth]{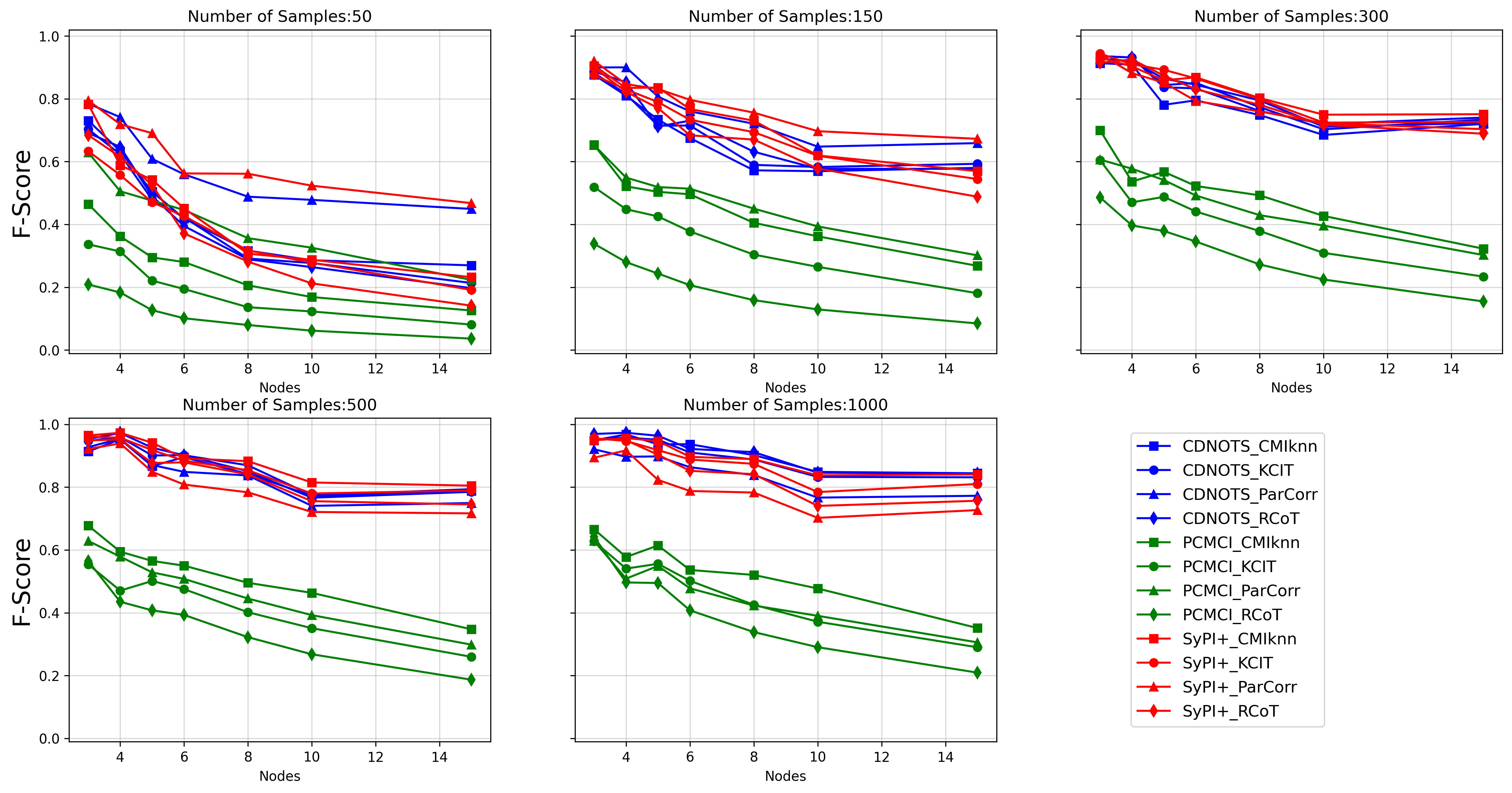}}
    \caption{F-score evaluation for different 
 methods}\label{fig:ci_fscore_3d}
\end{minipage}
\end{figure}

\begin{itemize}
    \item \textbf{Data Points Impact}: As expected, F-scores improve with more data points.
    \item \textbf{ParCorr Performance}: Excels in low data regimes.
    \item \textbf{KCIT Variants}: The HBE variant outperforms the SW variant, except with many nodes and only 50 data points (we just plotted HBE results).
    \item \textbf{RCoT Variants}: Little difference between SW and HBE variants.
    \item \textbf{RCoT vs. KCIT}: RCoT is faster for large datasets on a CPU. However, with GPU implementations, KCIT's deterministic nature makes it preferable.
    \item \textbf{CMIknn}: It is the most accurate in mid range data regimes, but its usage in PCMCI and CDNOTS in large datasets is very slow.
\end{itemize}

\textbf{Structural Hamming Distance (SHD)}: 
\myref{Figure}{fig:shd_3d}, displays SHD for each CI test across different node counts and data points. as can be observed SyPI+ discovered graphs are closer to ground truth in small and mid-sized graphs, however, in larger graphs CDNOTS perform slightly better.

\begin{figure}[!bt]
\begin{minipage}{\linewidth}
    \centerline{\includegraphics[width=\linewidth]{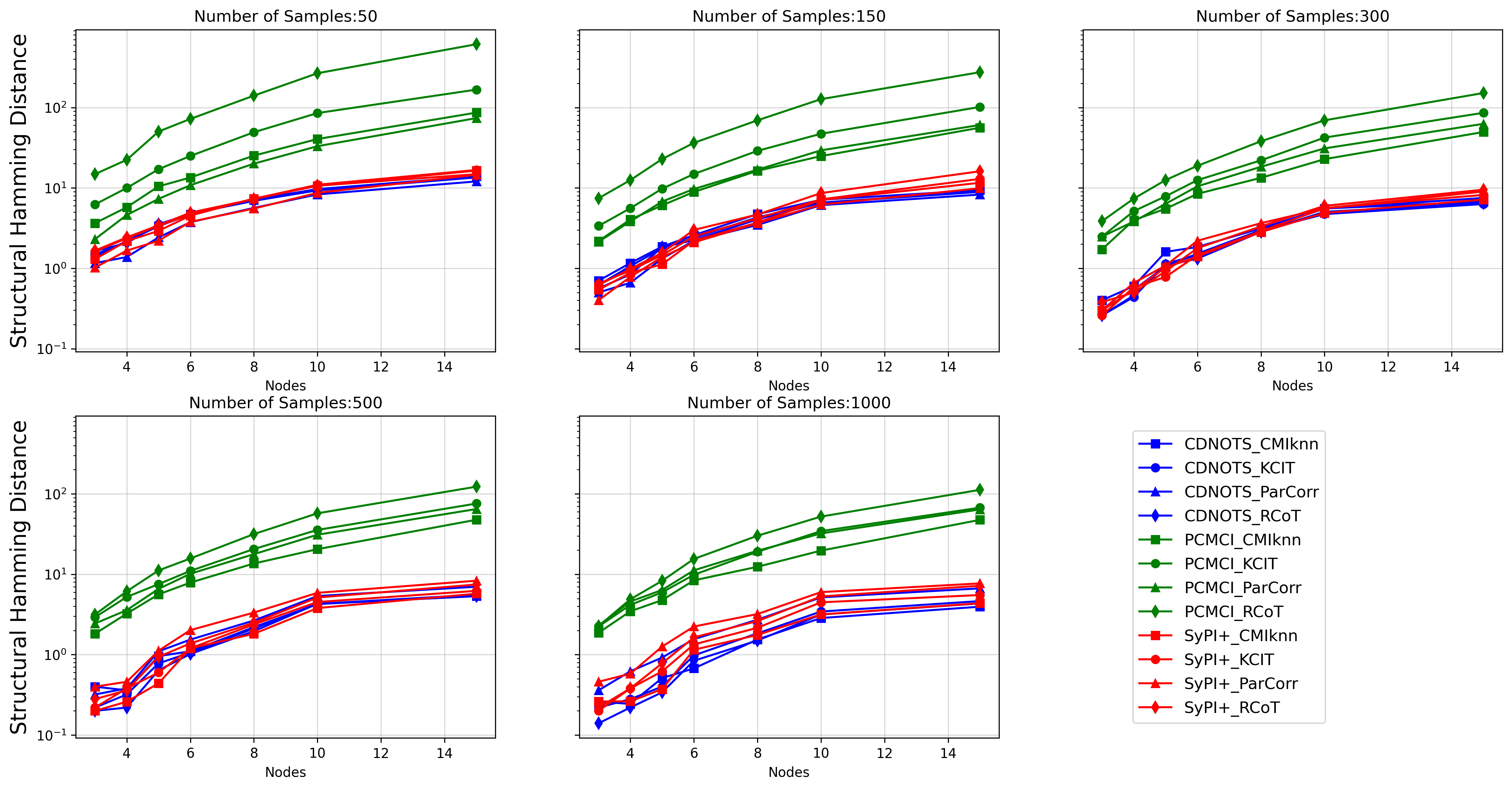}}
    \caption{SHD for different 
 methods}\label{fig:shd_3d}
\end{minipage}
\end{figure}

\section{Case Study: Causal Connections Between Major Banks}\label{sec:case_study} 

In this case study, we aim to uncover causal relationships between major Wall Street banks using weekly Credit Default Swap (CDS) data from end of January 2004 to September 2008, just before the collapse of Lehman Brothers. The central question we seek to address is: \textit{Should Lehman Brothers have been saved?}

\subsection{Data Preparation}

The CDS data spans from January 2004 to September 2008, covering the critical period leading up to the financial crisis. The banks included in our study are Lehman Brothers, Goldman Sachs, Morgan Stanley, AIG, UBS, Wells Fargo, JPMorgan Chase, Deutsche Bank, Barclays, Citigroup, and Bank of America. Each bank's weekly CDS spread is used as the primary data source for our analysis. To ensure that our analysis focuses on the specific interactions between these banks, we first conducted a series of data preparations:

\textbf{Systematic Factors}: Initial exploratory analysis of the data showed all banks are highly correlated with each other, which can be an indicator of common confounding factors that affect all banks. To isolate the individual relationships between them, we removed systematic market factors that could influence all banks similarly. Specifically, we adjusted the CDS spreads by removing the influences of the VFH (ETF representing financial sector), the VIX Index (representing market volatility), IBOXUMAE Index (The Markit CDX North America Investment Grade Index), and NFCINTED Index (TED spread).\newline
    
\textbf{Residual Calculation}: We used Random Forest method to adjust for systematic factors, we calculated the residuals of the CDS spreads for each bank. These residuals represent the portion of the CDS spreads that cannot be explained by the common factors we considered above, allowing us to focus on the unique interactions between the banks.

It is important to acknowledge that there may be other factors not included in our data that systematically affect some or all banks. This can lead to spurious links in the discovered network. Since this case study is intended to demonstrate a potential application, we do not claim that all relevant factors have been considered, and thus we cannot guarantee the accuracy of the discovered links. Practitioners should thoroughly analyze their specific problem and incorporate all relevant factors before applying any causal discovery algorithm.

\subsection{Results and Analysis}
With the residuals prepared, we applied SyPI+ to uncover potential causal links between the banks. The data samples are limited (242 weeks) for this problem, using the findings from synthetic tests, we decided to use partial correlation as CI test, as it has shown good performance even for nonlinear cases in low data regimes. 

\begin{figure}[!bt]
\begin{minipage}{\linewidth}
    \centerline{\includegraphics[width=.6\linewidth]{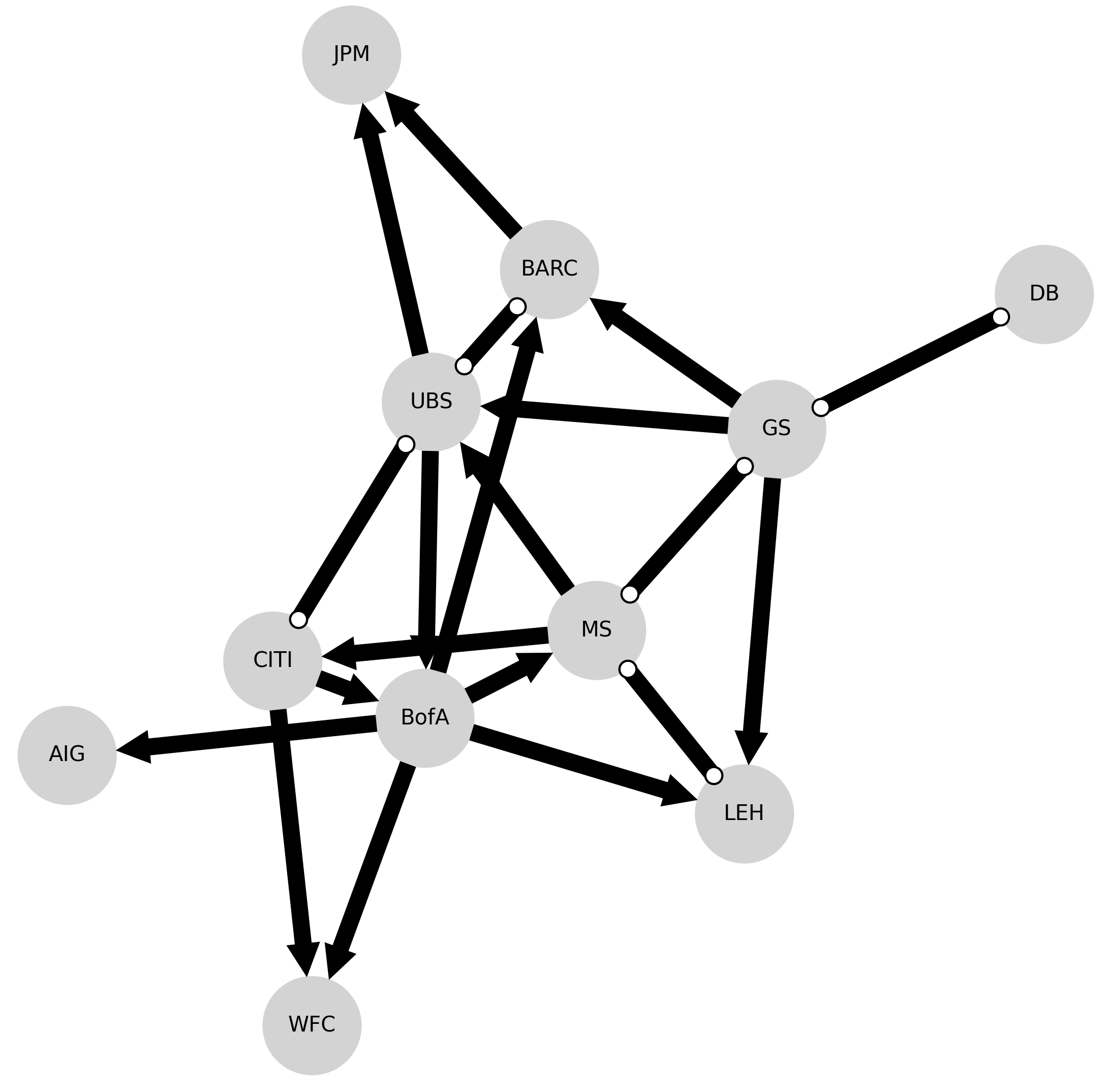}}
    \caption{CDS network for major banks, just before the collapse of Lehman Brothers}\label{fig:cds_banks}
\end{minipage}
\end{figure}

\myref{Figure}{fig:cds_banks}, displays default network learned from residual CDS data. Our analysis revealed both direct and indirect causal relationships among the banks. Notably, Lehman Brothers demonstrated strong causal links to Goldman Sachs, Morgan Stanley, and Bank of America, indicating that systemic stress originating from Lehman could have influenced these institutions and contributed to the broader intensification of the financial crisis.

\section{Conclusions}
In this study, we proposed a novel method to identify nonlinear relationships in autoregressive time series data. Our method requires significantly fewer conditional independence tests compared to existing methods. Reducing the number of CI tests results in greater statistical power leading to more robust performance. This is made evident in our synthetic experiments, where our method outperforms its competitors in low data scenarios as it identifies fewer false positive links. We also showed a potential use case of causal discovery by analyzing the default network constructed from CDS data, which can be useful to policy makers. Future work will extend our proposed method to account for non-stationarity in the data.

\section*{Acknowledgments}
We would like to express our sincere gratitude to Arun Verma and Abhinav Havaldar for their valuable suggestions and insightful discussions throughout the course of this work. Their expertise and thoughtful inputs have greatly enriched the content and clarity of this paper.

% \bibliographystyle{plainnat}
% % \bibliographystyle{abbrvnat}
% \bibliography{sample}

% \addbibresource{sample.bib}

\newpage
\appendix
\printbibliography
\end{document}